\documentclass[10pt,twocolumn,letterpaper]{article}

\usepackage{wacv}
\usepackage{times}
\usepackage{epsfig}
\usepackage{graphicx}
\usepackage{amsmath}
\usepackage{amssymb}
\usepackage{cite}
\usepackage{color}
\usepackage{booktabs}
\usepackage{caption}
\usepackage{array}
\usepackage{verbatim}
\usepackage{graphicx}\graphicspath{{Figures/}{tiff/}}
\usepackage{textcomp}
\usepackage{multirow}
\usepackage{url}
\usepackage{xspace}
\makeatletter
\let\@fnsymbol\@arabic
\makeatother

\wacvfinalcopy 

\ifwacvfinal\fi
\begin{document}

\title{Complex Event Recognition from Images with Few Training Examples}

\author{
    Unaiza Ahsan$^{*}$\\
    \texttt{\small uahsan3@gatech.edu}
  \and
	Chen Sun$^{**}$\\
    \texttt{\small chensun@google.com}
  \and
	James Hays$^{*}$\\
	\texttt{\small hays@gatech.edu}
  \and
	Irfan Essa$^{*}$\\
    \texttt{\small irfan@cc.gatech.edu}
  \and
    {\small *Georgia Institute of Technology}
  \\
    {\small **University of Southern California\thanks{The author currently works at Google.}}
}

\maketitle
\ifwacvfinal\thispagestyle{empty}\fi

\begin{abstract}
   We propose to leverage concept-level representations for complex event recognition in photographs given limited training examples. We introduce a novel framework to discover event concept attributes from the web and use that to extract semantic features from images and classify them into social event categories with few training examples. Discovered concepts include a variety of objects, scenes, actions and event sub-types, leading to a discriminative and compact representation for event images. Web images are obtained for each discovered event concept and we use (pretrained) CNN features to train concept classifiers. Extensive experiments on challenging event datasets demonstrate that our proposed method outperforms several baselines using deep CNN features directly in classifying images into events with limited training examples. We also demonstrate that our method achieves the best overall accuracy on a dataset with unseen event categories using a single training example. 
\end{abstract}

\section{Introduction}

\begin{figure}[!t]
\begin{center}
   \includegraphics[width=.8\linewidth]{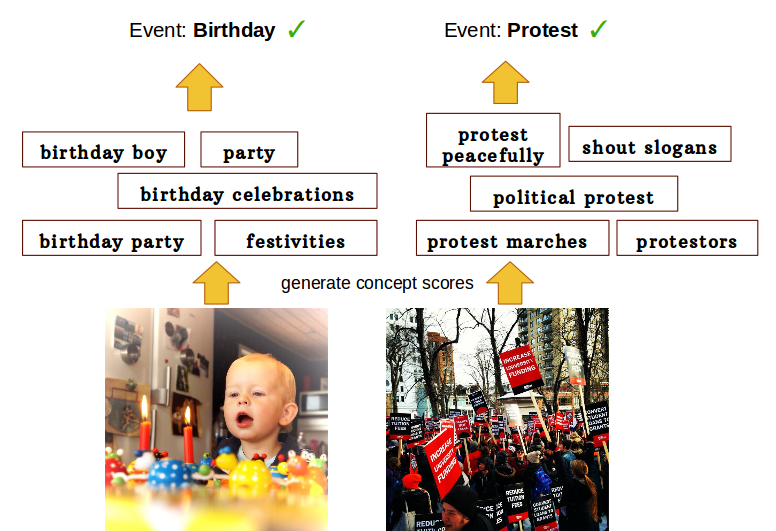}
\end{center}
\captionsetup{justification=justified}
   \caption{Event concepts as an intermediate feature representation for recognizing social events in photographs.}
\label{fig:f1}
\end{figure}

The widespread adoption of smart-phones coupled with easy to use photo sharing services has resulted in a massive increase in images shared online. A large number of these images are snapshots of special occasions, which we refer to as social events such as birthdays, graduations, weddings \etc or news events such as political campaigns, natural disasters or marathons. Many of these images do not have clean textual labels hence identifying events from visual content alone is non-trivial. The recent success of deep Convolutional Neural Networks (CNNs) in object and scene recognition has resulted due to large labeled training databases such as ImageNet \cite{ILSVRC15} and Places \cite{zhou2014learning}. Current approaches which use pretrained CNNs and fine-tune on datasets also require significant number of labeled examples. Since creating huge labeled datasets from the constantly evolving space of events is not realistic, we propose to learn an event concept-based representation and leverage that to identify rare events. Discovering web-driven concepts using Wikipedia and Flickr tags, we aim to categorize social events from static photographs when few labeled examples are available. Images of social events inherently consist of a combination of objects (\eg `banner'), scenes (\eg `ground'), actions (\eg `shouting slogans'), event subtypes (\eg `speech') and attributes (\eg `protest peacefully'). Object appearance significantly changes when combined with different objects, actions and attributes in cluttered backgrounds. Hence recognizing events from static images requires us to explicitly learn concept classifiers, each concept a combination of objects, scenes, actions and attributes. We call these \textit{event concepts} (see Figure \ref{fig:f1}). 

Event recognition approaches that use concepts or attributes have previously been applied to video-based events \cite{liu2013video, ye2015eventnet, singh2015selecting, ma2013complex} where temporal dynamics play an important role in recognizing what is happening in the video. This makes event recognition from a single photograph an interesting and challenging problem domain. Several attribute-based recognition methods require datasets annotated with all the concepts \cite{lampert2009learning, kumar2009attribute} which is a tedious process. To bypass the need for manual concept labeling and inspired by the recent `webly supervised' learning approaches \cite{chen2013neil, divvala2014learning, chen2015webly} that use web content to discover visual concepts, we propose an event concept learning framework using Wikipedia to generate event categories and Flickr tags as our initial pool of concepts. From noisy Flickr tags, we generate segments or phrases using a tweet segmentation algorithm proposed by Li \etal \cite{li2012twevent} which is a method designed specifically to extract event-centric phrases from noisy twitter streams. Finally, we project each event category on to a word embedding pretrained on the Google News Dataset using the popular \textsl{word2vec}~\cite{mikolov2013efficient} approach, extract nearest neighbors and add them to the pool of segmented phrases. We extract images related to each concept from MS Bing image search engine and compute deep CNN \cite{le1990handwritten} features extracted from a pretrained network on all the images and train concept classifiers. The concept scores predicted on a given test image form the final features for event images. 

Our \textbf{primary contributions} are:

\begin{enumerate}\itemsep 2pt \parskip 0pt \parsep 0pt
 \item A novel framework which involves using web data to discover event related concepts and employing efficient concept pruning strategies that result in clean, relevant and diverse event concepts. 
 \item A concept-based representation that not only improves single-shot event classification performance but can also be generalized to those categories which were not used during concept discovery.
 \item A large scale Social Event Image Dataset (SocEID) comprising 37,000 general event images belonging to 8 event categories as well as a challenging Rare Events Dataset (RED) comprising 7,000 images belonging to 21 specific real world events. 
\end{enumerate}

\section{Related Work}
Attributes have been used to describe objects (both fixed \cite{lampert2009learning, farhadi2009describing, chen2012describing, welinder2010caltech} and relative \cite{parikh2011relative}), faces \cite{kumar2009attribute}, scenes \cite{patterson2012sun} and actions \cite{liu2011recognizing, fu2012attribute}. These attribute detectors are then run on new images for high level recognition \cite{vogel2007semantic, wang2009learning}. Researchers have explored creating a set (or bank) of detectors pretrained on objects such as Object Banks \cite{li2010object}, an ontology of abstract concepts such as Classemes \cite{torresani2010efficient} or scene attributes \cite{patterson2012sun, ciocca2011halfway}. 

Another line of work proposes learning visual concepts from the Web with minimal human supervision (`webly supervised approaches'). NEIL \cite{chen2013neil} uses image search engine results in a semi-supervised setting to learn and train visual concept detectors. LEVAN \cite{divvala2014learning} uses Google NGram corpus to extract all possible words related to a given concept, extracts images from image search engine and learn visual concepts related to any given keyword. The authors of \cite{li2013harvesting} use a multiple instance learning approach to learn concepts from image search results. Some approaches learn concepts from images and their labels \cite{zhou2015conceptlearner}, from image descriptions \cite{sun2015automatic} or by using a deep network \cite{chen2015webly} using principles from curriculum learning. Our proposed work is inspired by web supervision but for a different domain. The key difference between our approach and other webly supervised concept learning approaches is that our methods are designed to obtain \textit{event specific} concepts. We explain further in Section~\ref{approach}.

The earliest work addressing event classification from static images \cite{li2007and} classifies sports events (rowing, rock climbing \etc) using object and scene information. Some related approaches \cite{jain2008selective, bossard2013event} require scene geometry or temporal alignment between event images to identify individual events. Recently \cite{xiong2015recognize} propose to train two deep networks; one on images and the second one on spatial maps of detected people/objects at different scales for event recognition. Our work aims to learn relevant concepts from the web and uses pretrained CNNs for feature extraction thus saving training time. More importantly, training a deep network for identifying events in images requires a large labeled dataset. We attempt to eliminate that requirement by discovering and using general event concepts from the web. 

Motivated by the need to learn with few labeled examples, vision researchers have addressed one-shot learning to learn object classifiers \cite{miller2000learning, bart2005cross, fei2006one, lake2011one, lake2013one} and more recently used deep networks \cite{koch2015siamese, held2015deep} and part-based models \cite{wong2015one}. Video event recognition community suffers from lack of labeled data for training too. Hence for recognizing events from few examples, many approaches use concepts as intermediate representations for event recognition \cite{wu2014zero, liu2013video}. Ma \etal \cite{ma2013complex} use labels in external videos as concepts and jointly model concept classification and event detection. Chen \etal \cite{chen2014event} use Flickr tags to discover concepts for an event and its associated text description. Cui \etal \cite{cui2014building} propose Concept Bank which consists of events mined from WikiHow and concepts from Flickr tags. Ye \etal \cite{ye2015eventnet} extend the previous work and propose to arrange events and their concepts in a hierarchy learned from WikiHow articles and YouTube descriptions. Shao \etal \cite{shao2015deeply} generate video event attributes using crowd-based annotation and use motion channels and appearance to train a deep model. Yang \etal \cite{yang2015automatic} learn video concepts from YouTube descriptions and Flickr tags. They generate event concepts using a tweet segmentation algorithm along with other metrics and train multiple classifiers for a single concept. The major difference between their work and ours is that we begin from event labels instead of descriptions, target image-based event recognition when few labeled examples are available and integrate word2vec based concepts into our concept pool. 

\section{Approach}\label{approach}
We begin by asking the question, ``What if all events had limited labeled examples?'' In reality, there are several events for which labeled image datasets are available \eg birthdays and weddings. We develop methods and test our approach on datasets containing these popular events by taking only a single labeled example from each category as training data. The rest of the dataset is used for testing. This approach enables us to determine whether our proposed methods work on popular events before taking on the harder task of identifying rare events (using our RED Dataset) from images. 

Our proposed method is an adaptation of the webly supervised learning approaches to learn visual concepts relevant to specific event categories. Extracting all visual concepts related to a keyword such as `birthday' from the web using the approach of \cite{divvala2014learning} results in many concepts that have either little to do with a birthday event or is not generalizable to real world images. Those concepts include birthday settings advertised by event planners online and objects associated with birthday with a clean background and a canonical viewpoint \cite{mezuman2012learning}. Hence we argue for event-specific concepts for complex event recognition from images with few labeled examples. 

Our approach is divided into three main parts: Event Concept Discovery, Training Concept Classifiers and Prediction of Concept Scores for Event Classification. 

\begin{table}[t]
\centering
\resizebox{0.5\textwidth}{!}{%
\begin{tabular}{|l|l|l|}
\hline
\textbf{air shows‎} & \textbf{auto show} & \textbf{beauty pageants‎} \\ \hline
\textbf{all star games‎} & \textbf{ballet show} & \textbf{beer festivals‎} \\ \hline
\textbf{american football match} & \textbf{ballroom dance} & \textbf{birdwatching} \\ \hline
\textbf{annual protests‎} & \textbf{balls} & \textbf{black friday} \\ \hline
\textbf{art exhibition} & \textbf{barbecue} & \textbf{boating} \\ \hline
\textbf{arts festivals‎} & \textbf{baseball} & \textbf{bowling} \\ \hline
\textbf{astronomy events‎} & \textbf{basketball match} & \textbf{boxing} \\ \hline
\end{tabular}%
}
\captionsetup{justification=centering}
\caption{Sample events mined from Wikipedia}
\label{T0}
\end{table}

\subsection{Event Concept Discovery}
We use Wikipedia to mine a list of events from its category `Social Events.' This list contains general events (such as birthday) and specific events (such as royal wedding). We do \textit{not} include specific events in our initial events list as the aim is to build a concept bank that is applicable for all events. If we build a concept bank for royal wedding, concepts such as `Kate Middleton,' 
`Buckingham Palace' \etc would be too specific to apply to generic event images. Thus we end up with 150 generic social events (see Table~\ref{T0}). This list of events was mined from Wikipedia but filtered (to remove specific events) manually.

Each event category in our list is used to query Flickr and we obtain the first 200 image results. We collect the tags (a set of words describing the image) of those images in the form of captions and this forms the noisy caption pool from which we aim to generate meaningful concepts that can describe general social events. We empirically observe that by increasing the number of images from which we mine Flickr tags, the noise in the data increases hence we limit ourselves to 200 images per event category. 

\paragraph{Tag Segmentation: }
For the set of events $E = \{e_1, e_2, ... e_n\}$ where, for our work $n = 150$, we have a set of tags $T = \{t_1, t_2, ... t_N\}$, $N=200n$ in our case. Our goal is to generate consecutive and non-overlapping segments $S = \{ s_1, s_2, ...s_m\}$. These segments can be a single word or phrases. We obtain a set of segments $S_i=\{ s_1, s_2, s_3,...s_{m_i}\} \subset S$ for each tag $t_i \in T$, $i \in \{1,...,N\}$, by applying a tweet segmentation method \cite{li2012twiner} which can be modeled as an optimization,
\begin{equation}
\underset{s_1, s_2, s_3,...s_{m_i}}{\operatorname{argmax}} \mbox{$Stk(t_i)$} = \sum_{j=1}^{m_i} Stk(s_j), 
\end{equation}
where \mbox{$Stk(\cdot)$} is a function that computes the \textit{stickiness} of a segment. Stickiness measures the probability that a segment of text is a `named entity' (name of a person, place or object) in a large text corpus or a knowledge base like Wikipedia. The final score for each segment $s_j$ is given by
\begin{equation}\label{eq4}
\mbox{score}(s_j) = \mbox{$Stk(s_j)$} \cdot V_{\mbox{\tiny{flickr}}}(s_j).
\end{equation}

\begin{figure*}[!t]
\begin{center}
   \includegraphics[height=4.0cm,width=0.7\linewidth]{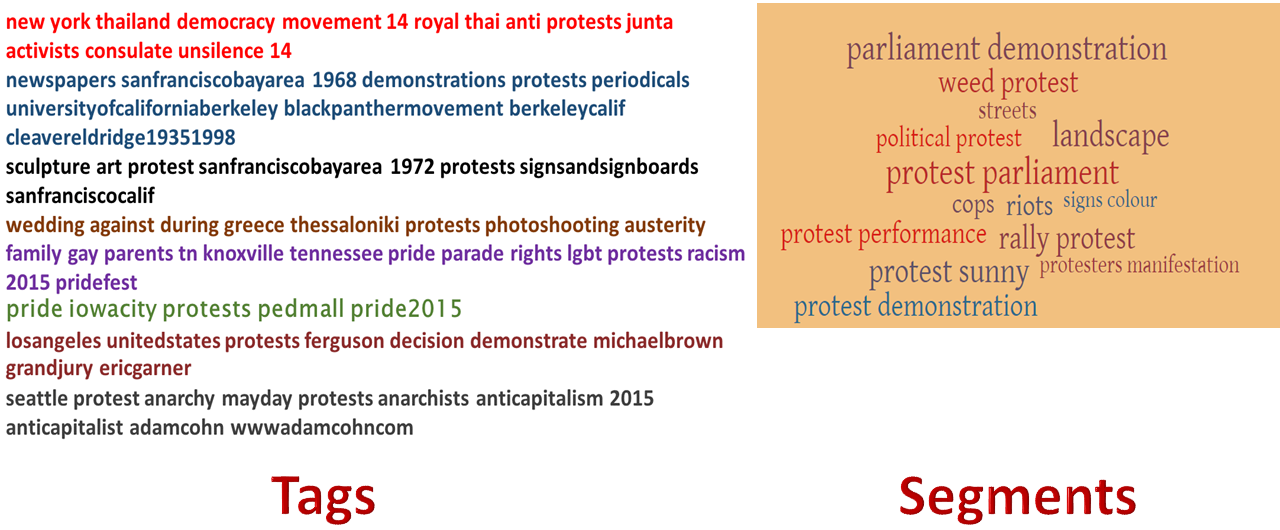}
\end{center}
\captionsetup{justification=justified}
   \caption{Generated segments from Flickr tags for event label `protest.'}
\label{fig:f3}
\end{figure*}

In Equation \ref{eq4} $\mbox{Stk}(s_j)$ refers to the stickiness score of the $jth$ segment and $V_{\mbox{\tiny{flickr}}}$ refers to \textit{visual representativeness} which is a measure of how visually coherent a word or phrase is. Words like `economy,' and `public,' if queried to an image search engine, result in ambiguous images, whereas a phrase like `birthday cake' generally returns very similar images. Hence it has a high visual representativeness score as compared to words like `economy' or `activity.' We obtain visual representativeness scores for each segment via a public dataset made available by Sun and Bhowmick \cite{sun2010quantifying} where they provide representativeness scores for the most popular tags on Flickr. After computing the final scores, we inspect the highest scoring segments to remove ambiguous or slang words. Figure \ref{fig:f3} shows some sample tags and the returned high scoring segments. Further details on tag segmentation can be found in \cite{li2012twiner}. 

\begin{table}[b!]
\centering
\resizebox{0.5\textwidth}{!}{%
\begin{tabular}{|l|l|}
\hline
dance & \begin{tabular}[c]{@{}l@{}}breakdancing, salsa dancing, argentinean tango, wows crowd, freerunning, reggae hiphop, \\ disco fever, flash mobs, street dance, dancers, breakbeat, hop, dance craze, dancefest, \\ bollywood bhangra, asian pop, dance workout, hip hop dance troupe, breakdancers\end{tabular} \\ \hline
\end{tabular}%
}
\captionsetup{justification=centering}
\caption{Nearest neighbors of the event `dance' in word2vec space}
\label{T0b}
\end{table}

Generating segments from tags is not enough to cover all aspects of a complex event. To expand the taxonomy and find event-specific concepts we additionally project each event label to word2vec space \cite{mikolov2013efficient}. We use an embedding trained on the Google News Dataset which consists of about 100 billion words. The model is available for public use (https://code.google.com/p/word2vec) and contains 300-dimensional vectors for 3 million words and phrases. The advantage of using Google News pretrained vectors is that we obtain semantically similar concepts for each event label. Table~\ref{T0b} shows the nearest neighbors of the event label `dance' in word2vec space.   

\begin{figure}[!t]
\begin{center}
   \includegraphics[height=4.5cm,width=0.9\linewidth]{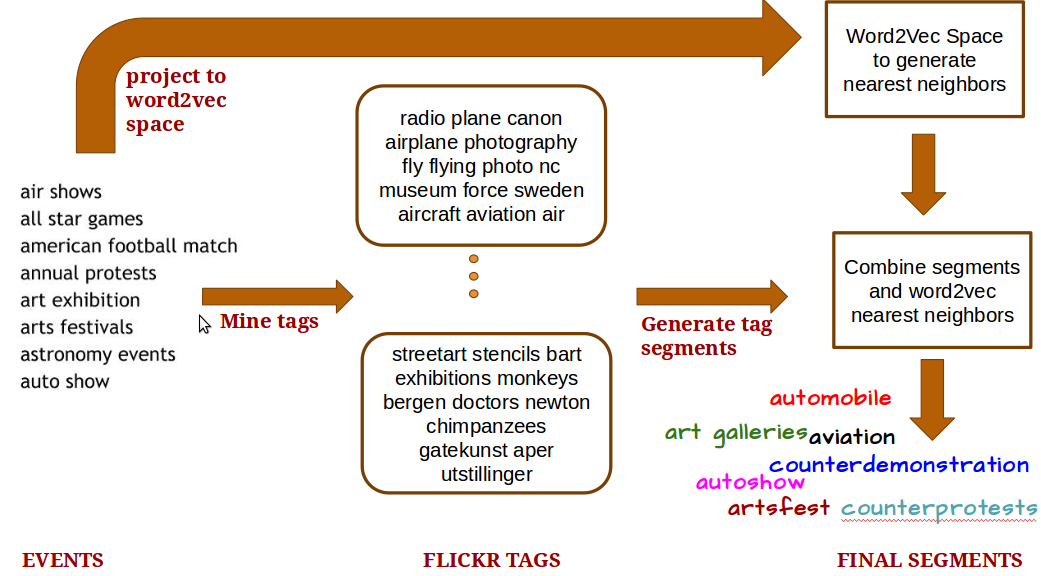}
\end{center}
\captionsetup{justification=justified}
  \caption{Event concept discovery pipeline for generic social events.}
\label{fig:f5}
\end{figure}

We extract top 20 nearest neighbors for each event label and add them to the tag segments generated via the segmentation scheme described above. This pool of event concepts is then filtered to remove duplicate concepts, slang words and foreign words. We finally end up with 856 event concepts. Our concepts not only include objects, scenes and actions but also include sub-events and their types. Our event concept discovery pipeline is shown in Figure \ref{fig:f5}.

\begin{figure}[!b]
\begin{center}
   \includegraphics[height=2.6cm,width=0.7\linewidth]{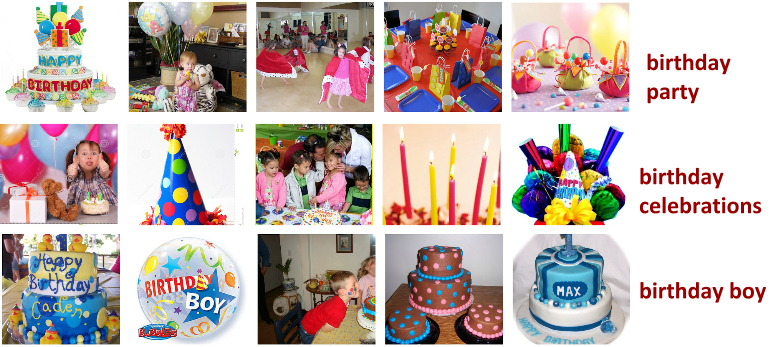}
\end{center}
\captionsetup{justification=justified}
   \caption{Examples of correlated event concepts}
\label{fig:f6}
\end{figure}


\begin{figure*}[!t]
\begin{center}
   \includegraphics[height=5.0cm,width=0.75\linewidth]{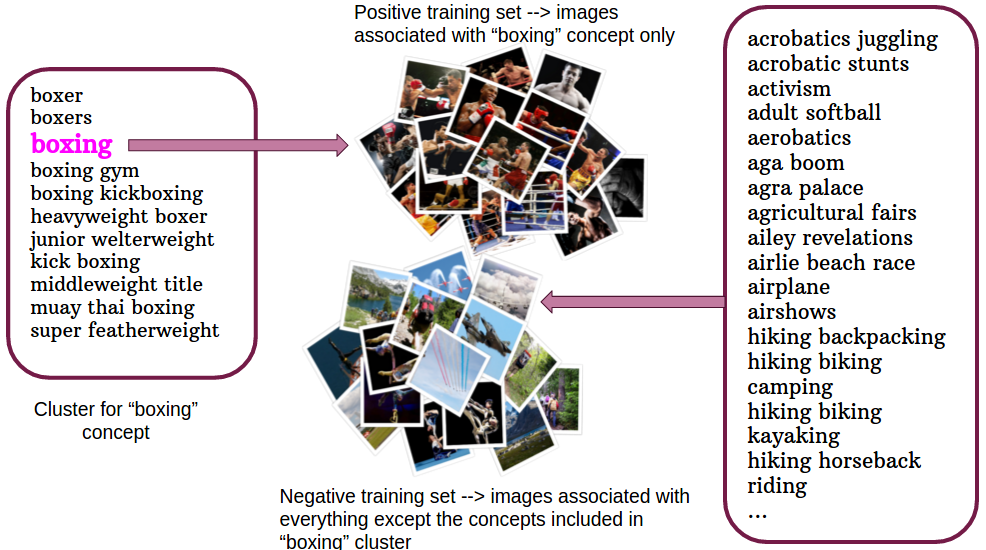}
\end{center}
\captionsetup{justification=justified}
   \caption{Selecting training images for `boxing' classifier}
\label{fig:f7}
\end{figure*}

\subsection{Training Concept Classifiers} \label{training_concepts}

Now we describe our approach for selecting training images for concept classifiers. Given a set of concepts $C = \big\{c_{1}, c_{2}, \dots c_{m}\big\}$, we input each concept as an image search query to Microsoft Bing and retrieve the top 100 images returned for each concept. Clipart, duplicate images and images containing only text are removed from the search results. Several concepts in $C$ are correlated with each other, resulting in similar images for different concepts. Training them independently without taking into account their correlation will lead to false negatives that will impact the concept classifier training negatively. The reason why we do not exclude correlated concepts in our concept pool $C$ is that they capture different (and thus important) aspects of a social event. For example, `birthday party,' `birthday boy' and `birthday celebrations' are three different concepts within the same event category (birthday) and they may have similar images. Thus when selecting training images for training the classifier for concept $c_{i}$, it is naive to sample negative training images from all concepts $c_{j}$ where $j \neq i$. Figure \ref{fig:f6} shows an example of correlated concepts and their associated images.

Hence, we first cluster all the concepts using their word2vec-based vector representations using minibatch k-means clustering \cite{sculley2010web}. 
We set $k = 150$. Thus for $ith$ concept $c_{i}$, belonging to a $y$-sized cluster, we obtain a list of concepts $C^{i}_{pos} = \big\{c_{i,pos_{1}}, 
c_{i,pos_{2}}, \dots c_{i,pos_{y}}\big\}$ that are highly correlated with the given concept $c_{i}$. We construct the concept classifier training set for concept $c_{i}$ as follows: 
\begin{itemize}
 \item Let $\xi^+ = \big\{I_{i}\big\}_{i=1}^u$ be the set of positive training images for classification where $u$ is the number of images retrieved for concept $c_{i}$. 
 \item Let $\xi^- = \big\{I_{j}\big\}_{j=1}^v$ be the set of negative training images for classification where $v$ is the number of images retrieved for the set 
of concepts $C^{i}_{neg} = \overline{C^{i}_{pos}} = C - C^{i}_{pos}$. 
\end{itemize}
In other words, we make sure that the set $\xi^-$ does not include images retrieved for any concept in the set $C^{i}_{pos}$ because those images are highly similar to the images in $\xi^+$ as they belong to the same cluster as $c_i$ (See Figure \ref{fig:f7}).

For each concept, we extract the CNN `fc7' layer activations as features from all its images, select the training and test examples as described above and input them to logistic regression classifiers. We select the classifier parameters through 5-fold cross validation and for all of our concept classifiers, the cross validation accuracy is above 90\%.

\subsection{Predicting Concept Scores for Classification}
After training all concept classifiers, we compute the classifier scores on images belonging to our evaluation datasets. For each image $I$, its feature vector is a concatenation of all concept classifier scores predicted on the image. Thus $f_I = \big\{x_{i}\big\}_{i=1}^m$ where $m$ is the total number of concepts and $x_i$ is the score predicted for $ith$ concept classifier. Finally, we use these features to classify event images into social events using a linear SVM with default parameters fixed for all experiments. We outline our experimental setup in detail in the next section. 

\section{Experiments and Evaluations}\label{experiment}
We evaluate our approach on four labeled event datasets with one-shot learning, that is, we use a single positive and negative training example from each class. We also report classification results for all-shot learning (using a 70-30 split) for comparison. 

\subsection{Datasets}
\begin{figure}[t]
\begin{center}
   \includegraphics[height=2.7cm,width=0.8\linewidth]{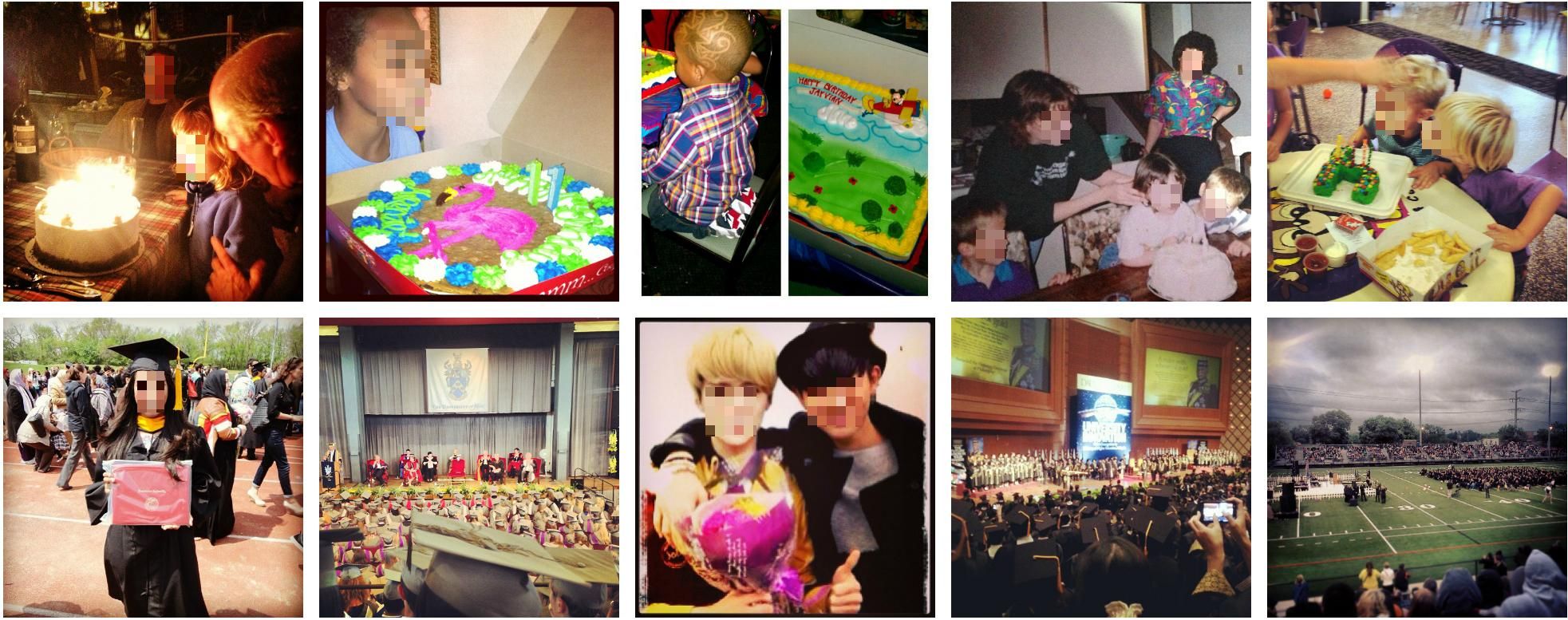}
\end{center}
\captionsetup{justification=justified}
   \caption{Sample images of the SocEID Dataset for two events: birthday (top) and graduation (bottom).}
\label{fig:f8}
\end{figure}

We evaluate our concept-based event recognition algorithm on the following four datasets:

\paragraph{1.~Social Event Image Dataset (SocEID):} This is the dataset we created in-house. We collected images of the following social events: birthdays, graduations, weddings, marathons/races, protests, parades, soccer matches and concerts. We queried Instagram and Flickr with a tag related to the event itself (`wedding day,' `Graduation 2014' \etc) and downloaded public images in chronological order determined by post date. Our dataset includes some relevant images from the NUS-WIDE dataset \cite{nus-wide-civr09} and the Social Event Classification subtask from MediaEval 2013 \cite{reuter2013social}. We passed all the images to 3 trusted coders (non-Turkers) and asked them to filter any image that did not depict a particular social event. We established the final scores on the images via majority vote and discarded all the rest. Finally, we ended up with nearly 37,000 images. Figure \ref{fig:f8} shows sample images from the SocEID dataset. 

\begin{figure*}[t!]
\begin{center}
   \includegraphics[height=5.0cm,width=0.8\linewidth]{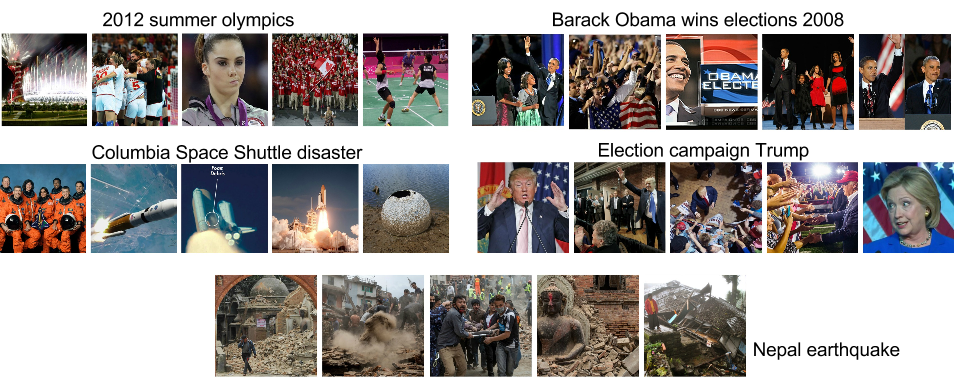}
\end{center}
\captionsetup{justification=justified}
   \caption{Sample images of the Rare Events Dataset.}
\label{fig:f8b}
\end{figure*}

\paragraph{2.~Web Image Dataset for Event Recognition (WIDER):} This dataset is introduced by \cite{xiong2015recognize} and consists of 50,574 images annotated with 61 classes.  

\paragraph{3.~UIUC Sports Event Dataset:} This dataset \cite{li2007and} consists of 1579 images belonging to 8 sports events categories such as badminton, bocce, sailing \etc 

\paragraph{4.~Rare Events Dataset (RED):} This is another in-house dataset we collected by querying MS Bing image search engine with a set of 26 `rare' event categories. We call them rare not on the basis of how frequently they occur in the world but on how seldom they are found in large labeled event image datasets. These event categories comprise of recent news events such as: \textit{Justin Trudeau elected}, \textit{election campaign Trump} and natural disasters such as \textit{Hurricane Katrina}, \textit{Hurricane Sandy} and \textit{Nepal earthquake} (see Figure ~\ref{fig:f8b}). The whole dataset comprises nearly 7,000 images and we do not remove any image from any event category manually. Note that these are all \textit{specific} events (the full list can be found in the supplementary section) and our main motivation behind collecting this dataset is twofold: i) Since few labeled examples are available for these events, it is a suitable test case for our claim that our learned concepts are a powerful intermediate representation to recognize events with few examples, ii) We want to test whether our discovered event concepts generalize to recognizing specific event images or not. 

\begin{figure*}[t!]
\begin{center}
   \includegraphics[height=4.0cm,width=0.5\linewidth]{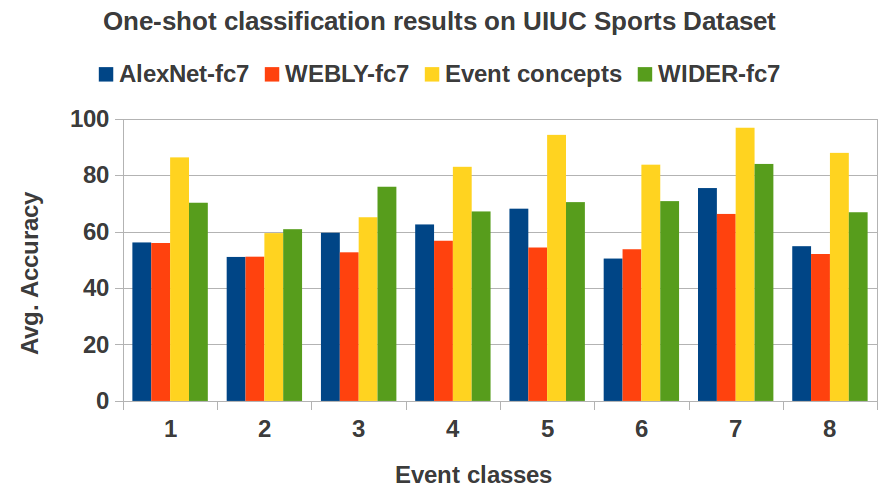}\includegraphics[height=4.0cm,width=0.5\linewidth]{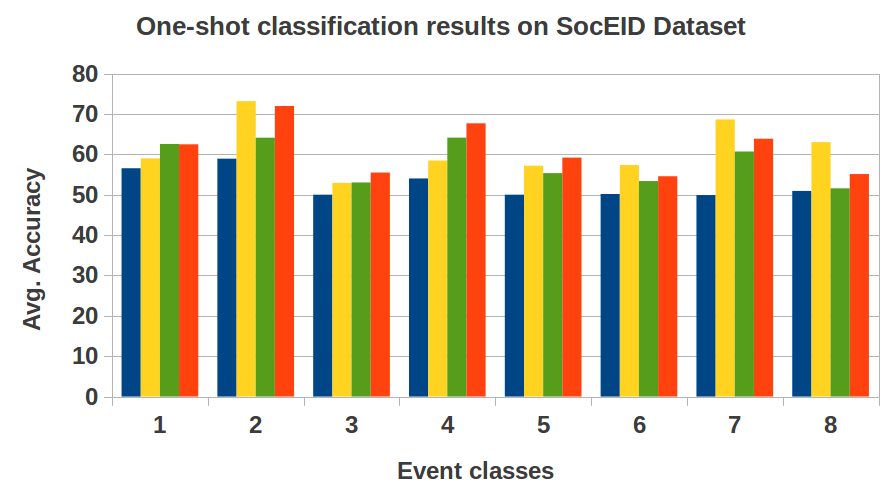}
\end{center}
\captionsetup{justification=justified}
   \caption{One-shot learning results on UIUC Sports Dataset and SocEID. }
\label{fig:f9}
\end{figure*}

\begin{table}[h!]
\centering
\resizebox{0.5\textwidth}{!}{%
\begin{tabular}{|l|l|}
\hline
rowing & \begin{tabular}[c]{@{}l@{}}rowing championships, rowing regatta, recreational boating, canoe trip, junior rowing, \\ canoe polo, standup pandling, swimming canoeing, recreational fishing, cable wakeboard\end{tabular} \\ \hline
polo & \begin{tabular}[c]{@{}l@{}}horseback riding, horse show, mountain biking, bronc riding, bareback bronc, \\ horseback ride, ranch rodeo, stampede rodeo, canoeing horseback, riding mountain\end{tabular} \\ \hline
\end{tabular}%
}
\captionsetup{justification=centering}
\caption{Top 10 predicted concepts for sports events `rowing' and `polo'}
\label{T2}
\end{table}

\begin{table*}[t!]
\centering
\resizebox{\textwidth}{!}{%
\begin{tabular}{@{}llllllllllllllll@{}}
\toprule
Features & 1 & 2 & 3 & 4 & 5 & 6 & 7 & 8 & 9 & 10 & 11 & 12 & 13 & 14 & 15 \\ \midrule
AlexNet-fc7 & 57.51 & 55.76 & 50.94 & \textbf{53.26} & 50.45 & \textbf{65.23} & 52.70 & 51.56 & 53.66 & 55.50 & 53.01 & 52.38 & 54.53 & \textbf{57.69} & 58.86 \\
WEBLY-fc7 & 57.97 & 55.54 & 53.05 & 53.07 & 57.43 & 64.59 & \textbf{54.32} & \textbf{54.26} & \textbf{55.46} & \textbf{60.19} & 56.03 & \textbf{60.07} & \textbf{55.13} & 56.31 & \textbf{60.70} \\
\textbf{Event concepts} & \textbf{60.46} & \textbf{56.62} & \textbf{63.64} & 52.16 & \textbf{59.63} & 64.96 & 51.71 & 51.59 & 54.80 & 56.87 & \textbf{57.82} & 59.29 & 49.00 & 57.05 & 59.47 \\\bottomrule
\end{tabular}%
}
\captionsetup{justification=centering}
\caption{Result of one-shot learning on WIDER Dataset}
\label{T3}
\end{table*}

\subsection{Experimental Setup}
We begin our experiments by training event concept classifiers. We have a total of 86,000 images associated with the concepts, retrieved from Microsoft Bing using the publicly available Bing crawler by Dengxin Dai.\footnote{\url{http://www.vision.ee.ethz.ch/~daid}} We use the Caffe \cite{jia2014caffe} deep learning framework to  extract CNN layer 7 activations (`fc7') as features for all the images using HybridCNN which is a publicly available CNN model pretrained on 978 object categories from ImageNet database \cite{ILSVRC15} and 205 scene categories from Places dataset \cite{zhou2014learning} using the AlexNet deep architechture \cite{krizhevsky2012imagenet}. For each concept, we select the positive and negative training features as described in Section \ref{training_concepts} and train L2-regularized logistic regression classifiers using the publicly available LIBLINEAR library \cite{fan2008liblinear}. Every image in our event datasets is input to each of the trained concept classifiers and a probabilistic concept score is computed on it. The fusing of concept scores form the final feature vector of that image. 
\subsubsection{Training with a Single Positive Image}
We conduct our one-shot learning experiment on the event datasets as follows: For event category $E$ with $P$ positive training features and $N$ negative training features we randomly sample a positive training feature $f^E_{p}$ and a negative training feature $f^E_{n}$. We concatenate the two and feed this into a binary linear SVM as training features. For testing, we simply take the rest of the positive features ($P - f^E_{p}$) and sample an equal number of negative features from the rest of the event categories in the dataset. Hence for all our experiments, the random baseline is 50\%. We run all experiments five times and average the per-class classification accuracies. 
\vspace{-4mm}
\subsubsection{Training with a 70\%-30\% split} 
In this experiment we take all of the labeled data into account and for each class, randomly select 70\% of images for training and test on the remaining images. This experiment shows the maximum accuracy our method can achieve given all of the training data available. It provides a nice comparison against the case where only a single labeled image is available for each class. We compare our results with several powerful baselines:

\begin{itemize}
 \item AlexNet \cite{krizhevsky2012imagenet} pretrained on ImageNet \cite{ILSVRC15} and Places \cite{zhou2014learning} databases, from which we extract 4096-dimensional layer fc7's activations and use them as features. We refer to this baseline as AlexNet-fc7 in the results. 
 \item Chen \etal \cite{chen2015webly} which is a recently proposed webly supervised CNN trained on about 2.1 million images downloaded from Google Images using popular vision datasets' labels as search queries. The authors use 2,240 objects, 89 attributes, and 874 scene labels from ImageNet \cite{ILSVRC15}, SUN database \cite{xiao2010sun} and NEIL knowledge base \cite{chen2013neil} and use principles from curriculum learning to train the network with easy examples first and then hard examples from Flickr. They show state of the art performance compared to AlexNet for objection detection and scene recognition. We use their `GoogleA' network which is trained on 2.1 million Google images. We refer to this baseline as WEBLY-fc7. 
 \item AlexNet \cite{krizhevsky2012imagenet} finetuned on WIDER database \cite{xiong2015recognize}. This baseline model is provided by the authors. We want to see how fc7 features extracted from this model perform with limited training data on our evaluation datasets (except WIDER) and whether our concept-level features are comparable to it. We refer to this baseline as WIDER-fc7. 
\end{itemize}

\begin{table*}[t!]
\centering
\resizebox{\textwidth}{!}{%
\begin{tabular}{llllllllllllllllllllllr}
\hline
Event classes & 1 & 2 & 3 & 4 & 5 & 6 & 7 & 8 & 9 & 10 & 11 & 12 & 13 & 14 & 15 & 16 & 17 & 18 & 19 & 20 & 21 & \multicolumn{1}{l}{Avg. Acc} \\ \hline
AlexNet-fc7 & 54.0 & 53.4 & 60.0 & \textbf{56.8} & 53.3 & 54.0 & 53.0 & 50.3 & 59.9 & \textbf{56.0} & 52.5 & 57.5 & 65.3 & 54.0 & \textbf{53.6} & 53.2 & 55.5 & 55.5 & \textbf{64.0} & 50.7 & 62.6 & 56.0 \\
WEBLY-fc7 & 52.2 & 58.1 & 60.9 & 56.2 & 53.9 & \textbf{56.3} & 50.4 & 55.5 & 67.8 & 54.9 & \textbf{61.0} & \textbf{61.3} & 59.6 & 55.6 & 47.8 & \textbf{61.0} & 53.2 & 52.1 & 56.0 & 52.6 & \textbf{69.0} & 57.0 \\
WIDER-fc7 & 50.4 & 55.5 & 57.2 & 52.2 & \textbf{55.8} & 55.1 & 51.1 & 52.8 & 55.5 & 52.6 & 52.6 & 58.9 & 63.3 & 53.7 & 47.3 & 49.8 & 53.4 & \textbf{55.6} & 61.3 & 52.8 & 59.3 & 54.6 \\ 
\textbf{Event concepts} & \textbf{57.5} & \textbf{58.9} & \textbf{67.7} & 55.5 & 54.8 & 53.1 & \textbf{53.9} & \textbf{58.6} & \textbf{75.1} & 54.2 & 60.8 & 55.4 & \textbf{73.4} & \textbf{56.2} & 52.0 & 58.0 & \textbf{56.2} & 52.4 & 58.7 & \textbf{53.3} & 64.8 & \textbf{58.6} \\
\hline
\end{tabular}%
}
\captionsetup{justification=centering}
\caption{Result of one-shot learning on RED Dataset}
\label{T4}
\end{table*}

\section{Results and Discussion}
Our one-shot learning result on UIUC Sports Events dataset (Figure~\ref{fig:f9} left) shows that the event concept features significantly outperform all the baselines in 6 out of 8 events. From 1-8, the UIUC Sports event categories are: badminton, bocce, croquet, polo, rock climbing, rowing, sailing and snowboarding. The main reason why this occurs is that our initial event list and hence our discovered event concepts include several sports events and their subtypes. For example, for all the images labeled with the event ``rowing'' and ``polo'' in the UIUC Sports Dataset, we count the top 10 most frequently predicted concepts. Table~\ref{T2} qualitatively shows that our method extracts relevant concepts consistently accross the set of UIUC Sports events images. 

Our one-shot learning experiment on the SocEID dataset (Figure~\ref{fig:f9} right) results in event concepts outperforming the baselines in 4 out of 8 categories. From 1-8, the categories are: birthday, concert, graduation, marathons, parade, protest, soccer and wedding. This is very likely due to the nature of images found in our dataset. Our dataset contains very clean images from the Web with significant visual cues of popular events such as birthdays. Thus the webly supervised network of \cite{chen2015webly} and WIDER \cite{xiong2015recognize} finetuned on event images is able to discriminate between the different event classes based on cues such as graduation caps and bridal gowns in the graduation and wedding pictures respectively even if limited training examples are present. 

Our experiments on the WIDER dataset \cite{xiong2015recognize} yield interesting insights. There are 61 classes in the dataset (the full list can be found in our supplementary material). For brevity, we only show the first 15 classes (we follow the order given by the authors). We test one-shot learning using our event concept features and compare them against the baselines, WEBLY-fc7 and AlexNet-fc7. Our event concept features outperform AlexNet-fc7 and WEBLY-fc7 in 31 classes when training with a single image. Table~\ref{T3} shows our results. From 1-15, the classes are: parade, handshaking, demonstration, riot, dancing, car-accident, funeral, cheering, election-campaign, press-conference, people-marching, meeting, group, interview and traffic. We note that the WIDER dataset contains not only events but also individual actions such as `cheering'. Our proposed approach uses concepts such as `cheering' to identify events which typically involve cheering such as dances, games or graduations (as the cheering concept will result in high probabilistic prediction on such events). However, the model is not trained to identify cheering alone. This can be verified by noting our performance scores in classes such as parade (1), demonstration (3), dancing (5) \etc. We score well on other categories (not shown in table) such as basketball, soccer, running, aerobics \etc Categories where our scores are comparable but not above the baselines are: sports coach trainer, greeting, surgeons, spa and stock market, to name a few. These categories are recognized more effectively by deep CNNs pretrained to recognize objects and scenes. 

Finally, we evaluate our method on the RED dataset (see Table~\ref{T4}) which is the most challenging because the images are highly diverse and consist of specific real world events. In one-shot learning on RED, for 10 out of 21 classes, our proposed method outperfoms the baselines. Thus our method is able to generalize to unseen event categories as our initial event list does not contain any of the rare event categories in the RED dataset. From left to right, categories are: `Russian airstrikes Syria,' `Boston Bombing,' `Nepal Earthquake,' `Arab Spring,' \etc (the full list can be found in the supplementary material).

\begin{table}[t!]
\centering
\resizebox{0.5\textwidth}{!}{%
\begin{tabular}{|l|l|l|l|l|}
\hline
 & \multicolumn{4}{c|}{Overall Average Accuracy (\%)} \\ \hline
Features & UIUC Sports & SocEID  & WIDER & RED \\ \hline
AlexNet-fc7 & 59.79 & 52.57 & 55.40 & 55.94 \\ \hline
WEBLY-fc7 & 55.39 & 60.89 & 58.14 & 56.92 \\ \hline
WIDER-fc7 & 70.81 & 58.11 & N/A & 54.58 \\ \hline
\textbf{Event concepts} & \textbf{82.09} & \textbf{62.98} & \textbf{58.29} & \textbf{58.59} \\ \hline
\end{tabular}%
}
\captionsetup{justification=centering}
\caption{Overall accuracy for one-shot learning on the evaluation datasets}
\label{T5}
\end{table}
The overall classification accuracies are shown in Table~\ref{T5}. For all the datasets, event image classification using a single training example with our proposed event concepts as features outperform the baselines which shows the strength of our approach to recognize complex real world events when limited labeled examples are available. 


We also evaluate our method against the baselines using all the available training data (70\%-30\% split). Table~\ref{T5} shows the overall classification accuracies on our evaluation datasets. Even when using all training examples, our event concept features are comparable to the state of the art in recognizing events from images. For two datasets, (UIUC Sports and WIDER) our method actually outperforms the state of the art. 

\begin{table}[t!]
\centering
\resizebox{0.5\textwidth}{!}{%
\begin{tabular}{|l|l|l|l|l|}
\hline
 & \multicolumn{4}{l|}{Overall Average Accuracy (\%)} \\ \hline
Features & UIUC Sports & SocEID & WIDER & RED \\ \hline
AlexNet-fc7 & 96.47 & \textbf{86.42} & 77.94 & 77.86 \\ \hline
WEBLY-fc7 & 95.16 & 83.66 & 77.85 & \textbf{79.39} \\ \hline
WIDER-fc7 & 93.85 & 80.42 & N/A & 76.64 \\ \hline
\textbf{Event concepts} & \textbf{96.68} & 85.39 & \textbf{78.59} & 77.57 \\ \hline
\end{tabular}%
}
\captionsetup{justification=centering}
\caption{Overall accuracy for all-shot learning on the evaluation datasets}
\label{T5}
\end{table}

\section{Conclusion}
In this paper we propose to discover event-specific concepts from the web to recognize complex events from images with few labeled examples. Our proposed framework discovers relevant concepts by combining segmented Flickr tags and word2vec nearest neighbors of event categories resulting in a compact intermediate representation which identifies real world events with only a single training example. We show the strength of our proposed method by evaluating on challenging datasets against powerful baselines which directly use CNN features pretrained on objects, scenes, attributes and events. It is interesting to note that in the problem domain of event recognition from visual content where only a few training examples are available, web-driven concept discovery and web images for training can result in highly discriminative intermediate representations which outperform directly using deep CNNs trained on millions of images and even deep CNNs finetuned on a large event dataset. 

{\small
\bibliographystyle{ieee}
\bibliography{egbib}
}

\end{document}